# Evolutionary Landscape and Management of Population Diversity

Maumita Bhattacharya
School of Computing & Mathematics
Charles Sturt University, Australia - 2640
mbhattacharya@csu.edu.au

**Abstract.** The search ability of an Evolutionary Algorithm (EA) depends on the variation among the individuals in the population [3, 4, 8]. Maintaining an optimal level of diversity in the EA population is imperative to ensure that progress of the EA search is unhindered by premature convergence to suboptimal solutions. Clearer understanding of the concept of population diversity, in the context of evolutionary search and premature convergence in particular, is the key to designing efficient EAs. To this end, this paper first presents a brief analysis of the EA population diversity issues. Next we present an investigation on a counter-niching EA technique [4] that introduces and maintains constructive diversity in the population. The proposed approach uses informed genetic operations to reach promising, but unexplored or under-explored areas of the search space, while discouraging premature local convergence. Simulation runs on a suite of standard benchmark test functions with Genetic Algorithm (GA) implementation shows promising results.

## 1 Introduction

Implementation of evolutionary algorithm (EA) requires preserving a population that maintains a degree of population diversity, while converging to a solution [7, 8, 9, 12, 13, 14, 15, and 16] in order to avoid premature convergence to sub-optimal solutions. It is difficult to precisely characterize the possible extent of premature convergence as it may occur in EA due to various reasons. The primary causes are algorithmic features like *high selection pressure* and *very high gene flow* among population members. Selection pressure pushes the evolutionary process to focus more and more on the already discovered better performing regions or "peaks" in the search space and as a result population diversity declines, gradually reaching a homogeneous state. On the other hand, unrestricted recombination results in high *gene flow* which spreads genetic material across the population, pushing it to a homogeneous state. Variation introduced through mutation is unlikely to be adequate to escape local optimum or optima [17]. While *premature convergence* [17] may be defined as the phenomenon of convergence to sub-optimal solutions, *gene-convergence* means loss of diversity in the process of evolution. Though, the convergence to a local or to the global optimum cannot necessarily be concluded from gene convergence, maintaining a certain degree of diversity is widely believed to help avoid entrapment in non-optimal solutions [3, 4].

In this paper we present an analysis on population diversity in the context of efficiency of evolutionary search. We then present an investigation on a counter niching-based EA that aims at combating gene-convergence (and premature convergence in turn) by employing intelligent introduction of constructive diversity [4].

The rest of the paper is organized as follows: Section 2 presents an analysis of diversity issues and the EA search process; Section 3 introduces the problem space for our proposed algorithm. Sections 4, 5 and 6 present the proposed algorithm, simulation details and discussions on the results respectively. Finally, Section 7 presents some concluding remarks.

## 2  Population Diversity and Evolutionary Search

The EA search process depends on the variation among the individuals or candidate solutions in the population. In case of genetic algorithm and similar EAs, the variation is introduced by the *recombination* operator combining existing solutions, and the *mutation* operator introducing noise by applying random variation to the individual's genome. However, as the algorithm progresses, loss of diversity or loss of genetic variation in the population results in low exploration, pushing the algorithm to converge prematurely to a local optimum or non-optimal solution. Exploration in this context means searching new regions in the solution space; whereas, exploitation means performing searches in the neighbourhoods which have been already visited. Success of the EA search process requires an optimal balance between exploitation and exploration.

In the context of EA, diversity may be described as the variation in the genetic material among individuals or candidate solutions in the EA population. This in turn may also mean variation in the fitness value of the individuals in the population. Two major roles played by population diversity in EA are as follows:

*Firstly*, diversity promotes exploration of the solution space to locate a single good solution by delaying convergence.

*Secondly*, diversity helps to locate multiple optima when more than one solution is present [8, 15 and 16].

Besides the role of diversity regarding premature convergence in static optimization problems, diversity also seems to be beneficial in non-stationary environments. If the genetic material in the population is too similar, i.e., has converged towards single points in the search space, all future individuals will be trapped at that single point even though the optimal solution has moved on to another location in the fitness landscape. However, if the population is diverse, the mechanism of recombination will continue to generate new candidate solutions making it possible for the EA to discover new optima.

The following sub-section presents an analysis of the impact of population diversity on premature convergence, based on the concepts presented in [13].

## 2.1 Effect of Population Diversity on Premature Convergence

Let $\vec{X} = (X_1, \ldots, X_N) \in S^N$ be a population of individuals $Y$ in the solution space $S^N$, where the population size is $N$; let $\vec{X}(0)$ be the initial population; **H** is a schema, i.e., a hyperplane of the solution space $S$. **H** may be represented by its defining components (defining *alleles*) and their corresponding values as $\mathbf{H}(a_{i1}, \ldots, a_{ik})$, where $K (1 \leq K \leq chromosome\ length)$. Leung et al. in [13] have proposed the following measures related to population diversity in canonical genetic algorithm.

*Degree of population diversity*, $\delta(\vec{X})$: Defined as the number of distinct components in the vector $\sum_{i=1}^{N} X_i$; and *Degree of population maturity*, $\mu(\vec{X})$: Described as $\mu(\vec{X}) = l - \delta(\vec{X})$ or the number of lost alleles.

With probability of mutation, $p(m) = 0$ and $\vec{X}(0) = \vec{X}_0$, according to Leung et al. [13] the following postulates hold true: For each solution, $Y \in \mathbf{H}\left(a_{i1}, \ldots, a_{i\mu(\vec{X}_0)}; \vec{X}_0\right)$, there exists a $n \geq 0$ such that *Probability* $\{Y \in \vec{X}(n) / \vec{X}(0) = \vec{X}_0\} > 0$. Conversely, for each solution, $Y \notin \mathbf{H}\left(a_{i1}, \ldots, a_{i\mu(\vec{X}_0)}; \vec{X}_0\right)$, and every $n \geq 0$ such that *Probability* $\{Y \in \vec{X}(n) / \vec{X}(0) = \vec{X}_0\} = 0$.

It is obvious from the above postulates that the search ability of a canonical genetic algorithm is confined to the minimum schema with $2^{\delta(\vec{X})}$ different individuals. Hence, the greater the degree of population diversity, $\delta(\vec{X})$, the greater is the search ability of the genetic algorithm. Conversely, a small degree of population diversity will mean limited search ability, reducing to zero search ability with $\delta(\vec{X}) = 0$.

## 2.2 Enhanced EAs To Combat Diversity Issues

No mechanism in a standard EA guarantees that the population will remain diverse throughout the run [17, 25]. Although there is a wide coverage of the fitness landscape at initialization due to the random initialization of individuals' genomes, selection quickly eliminates the least fit solutions, which implies that the population will converge towards similar points or even single points in the search space. Since the standard EA has limitations to maintain population diversity, several models have been proposed by the EA community which either maintain or reintroduce diversity in the EA population [1, 2, 4, 5, 6, 10, 11, 14 and 18]. The key researches can be broadly categorized as follows [15]:

1. Complex population structures to control gene flow, e.g., the diffusion model, the island model, the multinational EA and the religion model.
2. Specialized operators to control and assist the selection procedure, e.g., crowding, deterministic crowding, and sharing are believed to maintain diversity in the population.
3. Reintroduction of genetic material, e.g., random immigrants and mass extinction models are aimed at reintroduction of diversity in the population.
4. Dynamic Parameter Encoding (DPE), which dynamically resizes the available range of each parameter by expanding or reducing the search window.
5. Diversity guided or controlled genetic algorithms that use a diversity measure to assess and control the survival probability of individuals and the process of exploration and exploitation.

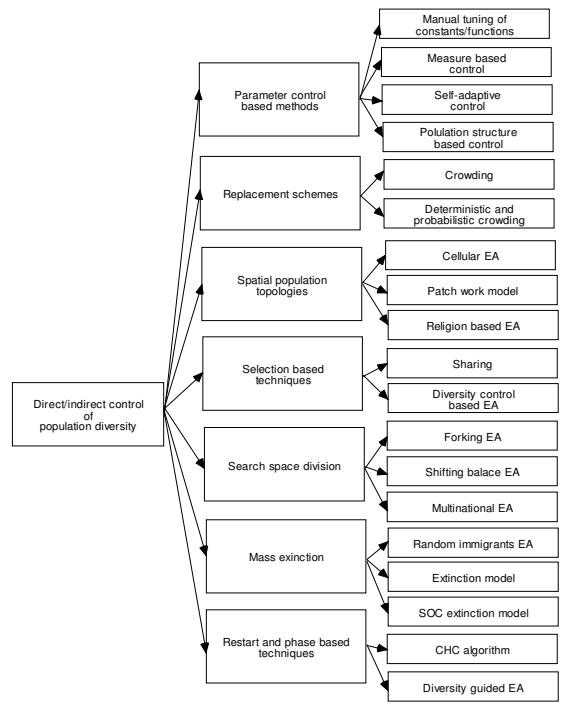

Fig 1. Direct or indirect control of population diversity in EA.

Figure 1 summarizes the major methods proposed to directly or indirectly control EA population diversity.

The Counter-Niching based EA framework presented in this paper, employs a synergistic hybrid mechanism that combines the benefits of *specialized operator* and *reintroduction of diversity*.

## 3   Understanding the Problem Space

Before we present our proposed approach, which aims at achieving constructive diversity, it is important to understand the problem space we are dealing with. For optimization problems the main challenge is often posed by the topology of the fitness landscape, in particular its ruggedness in terms of local optima. The target optimization problems for our approach are primarily multimodal. Genetic diversity of the population is particularly important in case of multimodal fitness landscape. Evolutionary algorithms are required to avoid and escape local optima or basins of attraction to reach the optimum in a multimodal fitness landscape.

Over the years, several new and enhanced EAs have been suggested to improve performance [1, 2, 4, 5, 6, 10, 11, 14, 20, 21, 22, 23 and 24]. The objectives of much of this research are twofold; *firstly*, to avoid stagnation in local optimum in order to find the global optimum; *secondly*, to locate multiple good solutions if the application requires so.

In the second case, i.e., to locate multiple good solutions, alternative and different solutions may have to be considered before accepting one final solution as the optimum. An algorithm that can keep track of multiple optima simultaneously should be able to find multiple optima in the same run by spreading out the search.

On the other hand, maintaining genetic diversity in the population can be particularly beneficial in the first case; the problem of entrapment in local optima. Mutation is not sufficient to escape local optima as selection traditionally favours the better fit solutions entrapped in local optima. Genetic diversity is crucial as a diverse population allows the recombination operators to find different and newer solutions.

*Remarks:* The issue is - *how much genetic diversity in the population is optimum*?

Unfortunately, the answer to the above question is not a straightforward one because of the complex interplay among the variation and the selection operators as well as the characteristics of the problem itself. Recombination in a fully converged population cannot produce solutions that are different from the parents; let alone better than the parents. Interestingly, Ishibuchi et al. [19] used a NSGA-II implementation to demonstrate that similar parents actually improved diversity without adversely influencing convergence. A very high diversity on the other hand, actually deteriorates performance of the recombination operator. Offspring generated combining two parents approaching two different peaks is likely to be placed somewhere between the two peaks; hindering the search process from reaching either of the peaks. This makes the recombination operator less efficient for fine-tuning the solutions to converge at the end of the run. Hence, the optimal level of diversity is somewhere between fully converged and highly diverse. Various diversity measures (such as Euclidean distance among candidate solutions, fitness distance and so on) may be used to analyze algorithms to evaluate their diversity maintaining capabilities.

In the following sections we investigate the functioning and performance of our proposed Counter Niching-based Evolutionary Algorithm [4].

## 4   Counter Niching EA: The Operational Framework

To attain the objective of introducing constructive diversity in the population, the proposed technique first extracts information about the population landscape before deciding on introduction of diversity through informed mutation. The aim is to identify locally converging regions or *donor* communities in the landscape whose redundant less fit members (or individuals) could be replaced by more promising members sampled in un-explored or under-explored sections of the decision space. The existence of such communities is purely based on the position and spread of individuals in the decision space at a given point in time. Once such regions are identified, random sampling is done on yet to be explored sections of the landscape. Best representatives found during such sampling, now replace the worst members of the identified *donor* regions. Best representatives are the ones that are fitness wise the fittest and spatially the farthest. Here, average Euclidean distance from representatives of all already considered regions (stored in a "memory" array) is the measure for spatial distance. Regular mutation and recombination takes place in the population as a whole. The basic framework is as depicted in Figure 2.

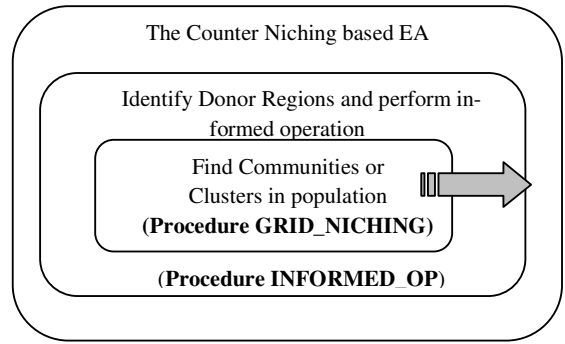

Fig 2. The COUNTER NICHING based EA framework

The task described in Figure 2 is carried out by the following three procedures:

1. **Procedure COUNTER NICHING EA:** This is the main algorithm that calls the procedures GRID_NICHING and INFORMED_OP. Basically, COUNTER_NICHING_EA has a very similar construct to a canonical genetic algorithm (see Figure 3) except that the genetic operations (recombination and mutation) are performed via procedures GRID_NICHING and INFORMED_OP. Procedure GRID_NICHING is used to identify the formation of clusters or locally genotypically converging regions in the solution space. Procedure INFORMED_OP, on the other hand, uses this clustering information to identify tendency towards fitness convergence, as this can be an early indication of premature convergence of the search process and hence, introduces diversity if necessary by a pseudo-mutation operator.

2. **Procedure GRID NICHING:** This function is called within COUNTER_NICHING_EA and is used to identify local genotypic convergence. Here, we have used the term niching simply to connote identification of environments of individuals in the population, based on their genotypical information. In other words, we try to identify roughly the individual clusters in the decision space based on their genotypic proximity. It may be noted that accuracy of the cluster boundaries is not of importance here. Instead, rough identification of cluster formation with reasonable amount of resources (runtime and memory space) is the prime objective.

   Thus the procedure GRID-NICHING, returns information about community or cluster formation in the population, for the current generation.

3. **Procedure INFORMED OP:** The procedure INFORMED_OP is second in order to be called by COUNTER_NICHING_EA. This function is used for performing the genetic operations (recombination and mutation) along with an informed mutation in appropriate cases. The INFORMED_OP algorithm searches for locally converging communities with too many members of similar fitness. To achieve this, the clusters or regions in the list of "identified regions with high density" returned by GRID_NICHING are analyzed for potential fitness convergence. Redundant members of the high density clusters or regions with low fitness standard deviation (victim regions) are picked for replacement by promising members from relatively un-explored or under-explored sections (virgin zones) of the solution space. The idea is to explore greater parts of the solution space at the expense of these so-called redundant or extra members. We call this process informed mutation. The potential replacements are generated by random sampling of the solution space. A potential replacement thus generated is picked as actual replacement if it has fitness higher than the average fitness of the victim region and if it is furthest from all cluster centers compared to other candidates of similar fitness.

   However, *informed mutation* as explained above, thus operates on selected regions or communities only. Regular mutation and recombination is performed as usual on the entire population.

Figure 3 presents the procedure COUNTER_NICHING_EA. For details on the procedures GRID NICHING and INFORMED OP, we refer to our previous work in [4].

## 5   Simulations

### 5.1   Test Functions

Following the standard practice in the evolutionary computation research community, we have tested the proposed algorithm on a set of commonly used benchmark test functions to validate its efficacy.

The benchmark test function set used in the simulation runs consists of minimization of seven analytical functions given in Table 1: Ackley's Path Function ($f_{ack}(x)$), Griewank's Function $f_{gri}(x)$, Rastrigin's Function $f_{rtg}(x)$, Generalized Rosen-brock's function $f_{ros}(x)$, Axis parallel Hyper-Ellipsoidal Function or Weighted Sphere Model $f_{elp}(x)$, Schwefel Function 1.2 $f_{sch-1.2}(x)$ and a rotated Rastrigin Function $f_{rrtg}(x)$.

Table 1. Description of Test Functions

| Function | Type | Global Minimum |
|---|---|---|
| $f_{ack}(x) = 20 + e - 20\exp\left(-0.2\sqrt{\frac{1}{n}\sum_{i=1}^{n} x_i^2}\right)$ $-\exp\left(\frac{1}{n}\sum_{i=1}^{n}\cos(2\pi \cdot x_i)\right)$ where $-30 \leq x_i \leq 30$ | Multimodal | $f_{ack}(x=0)=0$ |
| $f_{gri}(x) = \frac{1}{4000}\sum_{i=1}^{n}(x_i - 100)^2 -$ $\prod_{i=1}^{n}\cos\left(\frac{x_i - 100}{\sqrt{i}}\right) + 1$ where $-600 \leq x_i \leq 600$ | Multimodal, Medium epistasis | $f_{gri}(x=0)=0$ |
| $f_{rtg}(x) = \sum_{i=1}^{n}\left(x_i^2 - 10\cos(2\pi x_i) + 10\right)$ where $-5.12 \leq x_i \leq 5.12$ | Multimodal, No epistasis | $f_{rtg}(x=0)=0$ |
| $f_{ros}(x) = \sum_{i=1}^{n-1}\left(100(x_{i+1} - x_i^2)^2 + (x_i - 1)^2\right)$ where $-100 \leq x_i \leq 100$ | Unimodal, High epistasis | $f_{ros}(x=1)=0$ |
| $f_{elp}(x) = \sum_{i=1}^{M} i x_i^2$ where $-5.12 \leq x_i \leq 5.12$ | Unimodal | $f_{elp}(x=0)=0$ |
| $f_{sch-1.2}(x) = \sum_{i=1}^{M}\left(\sum_{k=1}^{i} x_k\right)^2$ where $-564 \leq x_i \leq 64$ | Unimodal, High epistasis | $f_{sch-1.2}(x=0)=0$ |
| $f_{rrtg}(x) = 10M + \sum_{i=1}^{M}\left(y_i^2 - 10\cos(2\pi y_i)\right)$ where $y = Ax$ with $A_{i,i} = 4/5$, $A_{i,i+1} = 3/5$ (i odd), $A_{i,i-1} = -3/5$ (i even), $A_{i,k} = 0$ (the rest) | Multimodal | $f_{rrtg}(x=0)=0$ |

Schwefel's function 1.2 and Rosenbrock's function are unimodal functions, but they have a strong epistasis among their variables. Griewank's function has very small but numerous minima around the global minimum, although it has a unimodal shape on a large scale. Rastrigin's function also has many local minima. However, it has no epistasis among its variables.

### 5.2 Algorithms Considered For Comparison

The algorithms used in the comparison are as follows:

1. The "standard EA" (SEA)
2. The self organized criticality EA (SOCEA)
3. The cellular EA (CEA), and
4. The diversity guided EA (DGEA)

The SEA uses Gaussian mutation with zero mean and variance $\sigma^2 = 1 + \sqrt{t+1}$. The SOCEA is a standard EA with non-fixed and non-decreasing variance $\sigma^2 = POW(10)$, where $POW(\alpha)$ is the power-law distribution. The purpose of the SOC-mutation operator is to introduce many small, some mid-sized, and a few large mutations. The effect of this simple extension is quite outstanding considering the effort to implement it in terms of lines of codes. The reader is referred to [15] for additional information on the SOCEA. Further, the CEA uses a 20x20 grid with wrapped edges. The grid size corresponds to the 400 individuals used in the other algorithms. The CEA uses Gaussian mutation with variance $\sigma^2 = POW(10)$, which allows comparison between the SOCEA and this version of the CEA. Mating is performed between the individual at a cell and a random neighbour from the four-neighbourhood. The offspring replaces the center individual if it has a better fitness than the center individual. Finally, the DGEA uses the Gaussian mutation operator with variance $\sigma^2 = POW(1)$. The diversity boundaries were set to $d_{low} = 5.10^{-6}$ and $d_{high} = 0.25$, which proved to be good settings in preliminary experiments.

**Algorithm 1:** Procedure COUNTER NICHING EA

```
 1: begin
 2: t ← 0
 3: Initialize population P(t)
 4: Evaluate population P(t)
 5: while (not<termination condition>)
 6: begin
 7:    t ← t+1
       (* Perform pseudo-niching of the population*)
 8:    Call Procedure GRID_NICHING
       (* Perform informed genetic operations *)
 9:    Call Procedure INFORMED_OP
 10:   Create new population using an elitist selection
         mechanism
 11: Evaluate P(t)
 14: end while
 15: end
```

Fig 3. The COUNTER NICHING based EA framework.

### 5.3 Experiment Set-up

Simulations were carried out to apply the proposed COUNTER NICHING based EA with real-valued encoding with parameters $N$ (population size) =300, $p_m$ (mutation probability) =0.01 and $p_r$ (recombination probability) =0.9. In case of the algorithms used for comparison as mentioned in Section 5.2, namely, (i) SEA (Standard EA), (ii) SOCEA (Self-organized criticality EA), (iii) CEA (The Cellular EA), and (iv) DGEA (Diversity guided EA), experiments were performed using real-valued encoding, a population size of 400 individuals, and binary tournament selection. Probability of mutating an entire genome was $p_m$ = 0.75 and probability for crossover was $p_r$ = 0.9. As mentioned in Section 5.2, CEA uses a 20x20 grid with wrapped edges, where the grid size corresponds to the population size of 400 individuals as used in the other algorithms. The compared algorithms all use variants of the standard Gaussian mutation operator. The algorithm uses an arithmetic crossover with one weight for each variable. All weights except one are randomly assigned to either 0 or 1. The remaining weight is set to a random value between 0 and 1.

All the test functions were considered in 20, 50 and 100 dimensions. Reported results were averaged over 30 independent runs, maximum number of generations in each run being only 500, as against 1000 generations in used [15] for the same set of test cases for the 20 dimensional scenarios. The comparison algorithms use 50 times the dimensionality of the test problems as the terminating generation number in general, while the COUNTER NICHING EA uses 500, 1000 and 2000 generations for the 20, 50 and 100 dimensional problem variants respectively.

All the simulation processes were executed using a Pentium® 4, 2.4GHz CPU processor.

## 6 Results and Discussions

This section presents the empirical results obtained by the COUNTER NICHING EA algorithm when tackling the seven test problems mentioned in Section 5.1 with dimensions 20, 50 and 100.

### 6.1 General Performance of COUNTER NICHING EA

Table 2 presents the error values, ($f(x) - f(x)^*$) where, $f(x)^*$ is the optimum. Each column corresponds to a test function. The error values have been presented for the three dimensions of the problems considered, namely 20, 50 and 100.

Table 2. Error values achieved on the test functions with simulation runs for COUNTER NICHING EA. Dimensions of each function considered are 20, 50 and 100.

| | | $f_{ack}(x)$ | $f_{gri}(x)$ | $f_{rtg}(x)$ | $f_{ros}(x)$ | $f_{elp}(x)$ | $f_{sch-1.2}(x)$ | $f_{rrtg}(x)$ |
|---|---|---|---|---|---|---|---|---|
| 20D | 1st (Best) | 1.00E-61 | 4.3E-62 | 1.11E-61 | 1.01E-60 | 2.01E-60 | 2.01E-50 | 3.89E-6 |
| | 7th | 1.11E-61 | 4.41E-62 | 1.131E-61 | 1.01E-60 | 2.01E-60 | 2.01E-50 | 3.89E-6 |
| | 15th (Median) | 1.11E-61 | 4.96E-62 | 1.210E-61 | 1.11E-60 | 2.89E-60 | 2.71E-50 | 3.9E-6 |
| | 22nd | 1.95E-61 | 5.61E-62 | 2.02E-61 | 1.92E-60 | 2.91E-60 | 2.91E-50 | 3.93E-6 |
| | 30th (Worst) | 3.11E-61 | 8.71E-62 | 3.30E-61 | 2.01E-60 | 2.91E-60 | 2.91E-50 | 3.99E-6 |
| | Mean | 1.12E-61 | 5.02E-62 | 1.21E-61 | 1.12E-60 | 2.92E-60 | 2.72E-50 | 3.9E-6 |
| | Std. | 8.33E-62 | 1.64E-62 | 8.6E-62 | 4.71E-61 | 4.62E-61 | 4.22E-51 | 3.88E-8 |
| 50D | 1st (Best) | 0.56E-29 | 1.00E-30 | 1.00E-30 | 1.21E-29 | 1.01E-30 | 2.21E-20 | 9.01 |
| | 7th | 0.71E-29 | 1.01E-30 | 1.01E-30 | 1.41E-29 | 1.01E-30 | 2.40E-20 | 9.01 |
| | 15th (Median) | 0.71E-29 | 1.01E-30 | 1.10E-30 | 1.90E-29 | 1.10E-30 | 2.90E-20 | 9.11 |
| | 22nd | 0.91E-29 | 1.91E-30 | 1.81E-30 | 1.92E-29 | 1.51E-30 | 2.91E-20 | 9.22 |
| | 30th (Worst) | 0.99E-29 | 1.99E-30 | 1.99E-30 | 1.98E-29 | 1.92E-30 | 2.91E-20 | 9.24 |
| | Mean | 0.73E-29 | 1.11E-30 | 1.11E-30 | 1.91E-29 | 1.11E-30 | 2.90E-20 | 9.12 |
| | Std. | 1.55E30 | 4.76E-31 | 4.41E-31 | 3.16E-30 | 3.66E-31 | 3.16E-21 | 0.098 |
| 100D | 1st (Best) | 1.00E-9 | 1.20E-9 | 1.90E-9 | 2.09E-9 | 2.09E-8 | 2.09E-5 | 10.52 |
| | 7th | 1.01E-9 | 1.51E-9 | 1.92E-9 | 2.91E-9 | 2.92E-8 | 2.59E-5 | 10.66 |
| | 15th (Median) | 1.12E-9 | 1.72E-9 | 1.99E-9 | 2.99E-9 | 2.99E-8 | 3.29E-5 | 11.09 |
| | 22nd | 1.36E-9 | 1.86E-9 | 2.21E-9 | 3.21E-9 | 3.21E-8 | 3.79E-5 | 11.61 |
| | 30th (Worst) | 1.36E-9 | 1.92E-9 | 2.92E-9 | 3.92E-9 | 3.90E-8 | 3.98E-5 | 11.79 |
| | Mean | 1.13E-9 | 1.81E-9 | 2.01E-9 | 3.03E-9 | 3.01E-8 | 3.69E-5 | 11.50 |

| | Std. | 1.61E-10 | 2.71E-10 | 3.88E-10 | 6.91E-10 | 5.81E-10 | 7.48E-6 | 0.5241 |

As each test problem was simulated over 30 independent runs, we have recorded results from each run and sorted the results in ascending order. Table 2 presents results from the representative runs: 1$^{st}$ (Best), 7$^{th}$, 15$^{th}$ (Median), 22$^{nd}$ and 30$^{th}$ (Worst), Mean and Standard Deviation (Std). The main performance measures used are the following:

**"A" Performance:** Mean performance or average of the best-fitness function found at the end of each run. (Represented as 'Mean' in Table 2).

**"SD" Performance:** Standard deviation performance. (Represented as 'Std.' in Table 2).

**"B" Performance:** Best of the fitness values averaged as mean performance. (Represented as 'Best' in Table 2).

As can be observed COUNTER NICHING EA has demonstrated descent performance in majority of the test cases. However, as can be seen from the highlighted segment (*highlighted in bold*) of Table 2, the proposed algorithm was not very efficient in handling the comparatively higher dimensional cases (50 and 100 dimensional cases in this example) for the rotated Rastrigin Function $f_{rrtg}(x)$. Keeping in mind the concept of *No Free Lunch Theorem*, this is acceptable as no single algorithm can be expected to perform favorably for all possible test cases. The chosen benchmark test functions represent a wide variety of test cases.

An algorithm's value can only be established if its performance is tested against that of existing algorithms for similar purposes. In the next phase of our experiments we have presented comparative performances of COUNTER NICHING EA as against SEA, SOCEA, CEA, and DGEA.

### 6.2 Comparative Performance of COUNTER NICHING EA

Simulation results obtained with COUNTER NICHING EA in comparison to SEA, SOCEA, CEA, and DGEA (see Section 5.2 for descriptions of these algorithms) are presented in Table 3. Results reported in this case, for COUNTER NICHING EA were averaged over 50 independent runs.

Table 3. Average fitness comparison for SEA, SOCEA, CEA, DGEA, and COUNTER NICHING EA[*]. Dimensions of each function considered are 20, 50 and 100. '-' appears where the corresponding data is not available.

| Function | SEA | SOCEA | CEA | DGEA | C_EA[*] |
|---|---|---|---|---|---|
| $f_{ack}(x)_{20D}$ | 2.494 | 0.633 | 0.239 | 3.36E-5 | 1.08E-61 |
| $f_{gri}(x)_{20D}$ | 1.171 | 0.930 | 0.642 | 7.88E-8 | 4.6E-62 |
| $f_{rtg}(x)_{20D}$ | 11.12 | 2.875 | 1.250 | 3.37E-8 | 1.21E-61 |
| $f_{ros}(x)_{20D}$ | 8292.32 | 406.490 | 149.056 | 8.127 | 1.0E-60 |
| $f_{elp}(x)_{20D}$ | - | - | - | - | 2.9E-60 |
| $f_{sch-1.2}(x)_{20D}$ | - | - | - | - | 2.7E-50 |
| $f_{rrtg}(x)_{20D}$ | - | - | - | - | 3.9E-6 |
| $f_{ack}(x)_{50D}$ | 2.870 | 1.525 | 0.651 | 2.52E-4 | 1.01E-29 |
| $f_{gri}(x)_{50D}$ | 1.616 | 1.147 | 1.032 | 1.19E-3 | 1.01E-30 |
| $f_{rtg}(x)_{50D}$ | 44.674 | 22.460 | 14.224 | 1.97E-6 | 2.01E-30 |
| $f_{ros}(x)_{50D}$ | 41425.674 | 4783.246 | 1160.078 | 59.789 | 1.91E-29 |
| $f_{elp}(x)_{50D}$ | - | - | - | - | 1.00E-30 |
| $f_{sch-1.2}(x)_{50D}$ | - | - | - | - | 2.9E-20 |
| $f_{rrtg}(x)_{50D}$ | - | - | - | - | 9.1 |
| $f_{ack}(x)_{100D}$ | 2.893 | 2.220 | 1.140 | 9.80E-4 | 1.00E-9 |
| $f_{gri}(x)_{100D}$ | 2.250 | 1.629 | 1.179 | 3.24E-3 | 1.80E-9 |
| $f_{rtg}(x)_{100D}$ | 106.212 | 86.364 | 58.380 | 6.56E-5 | 2.00E-9 |
| $f_{ros}(x)_{100D}$ | 91251.300 | 30427.63 | 6053.870 | 880.324 | 3.00E-9 |
| $f_{elp}(x)_{100D}$ | - | - | - | - | 2.99E-8 |
| $f_{sch-1.2}(x)_{100D}$ | - | - | - | - | 3.7E-5 |
| $f_{rrtg}(x)_{100D}$ | - | - | - | - | 11.51 |

These simulation results demonstrate COUNTER NICHING EA's superior performance as regards to solution precision in all the test cases, particularly for lower dimensional instances. This may be attributed to COUNTER NICHING EA's ability to strike a better balance between exploration and exploitation. However, the pro-posed algorithm's performance deteriorates with increasing dimensions. Also, the algorithm could not handle the high dimensional versions of the high epistatis rotated Rastrigin function to any satisfactory level. Table 4 depicts the runtimes for the tested algorithms for the 100 dimensional scenarios of four test cases used in our experiments. Considering the structures of the algorithms, a trade-off between solution accuracy and computational time can be expected for COUNTER NICHING EA. On the other hand, DGEA, which is designed to skip certain genetic operations depending on the level of population diversity, would be a clear winner in terms of computation time if all the algorithms are executed for the same number of generations in each run.

Table 4. Average runtime in milliseconds for SEA, SOCEA, CEA, DGEA and COUNTER_NICHING_EA[*] for the 100 dimensional scenarios. (Average of 100 runs with 2000 generations for COUNTER_NICHING_EA[*] and 5000 generations for other algorithms).

| Method | $f_{ack}(x)_{100D}$ | $f_{gri}(x)_{100D}$ | $f_{rtg}(x)_{100D}$ | $f_{ros}(x)_{100D}$ |
|---|---|---|---|---|
| SEA | 1128405 | 1171301 | 1124925 | 1087615 |
| SOCEA | 1528864 | 1562931 | 1513691 | 1496164 |
| CEA | 2951963 | 3656724 | 2897793 | 2183283 |
| DGEA | 864316 | 969683 | 819691 | 883811 |
| C_EA* | 418489 | 521800 | 491411 | 510266 |

For the reported results as shown in Table 3, the 100 dimensional scenarios of the test problems used 5000 generations for each of the compared algorithm, namely, SEA, SOCEA, CEA and DGEA. On the other hand, COUNTER NICHING EA used only 2000 generations to reach the reported results. Hence, for comparison purposes it is only fair to consider the computation time required by the different methods to reach comparable results. As can be observed from Table 4, despite its relatively complex algorithmic structure, COUNTER NICHING EA requires less computation time to reach better or comparable solution accuracy. We have also extended the simulation runs beyond the fixed number of generations and to the stagnation point. Here, stagnation point is defined by the generation with 500 successive generations of no fitness improvement preceding it. Table 5 summarizes the results for DGEA and COUNTER NICHING EA with fixed run and at stagnation. Both DGEA and COUNTER NICHING EA show some improvement over the results obtained with fixed number of generations in most cases. COUNTER NICHING EA still outperforms DGEA. Also, COUNTER NICHING EA has arrived at these superior results in much fewer generations. However, no significant improvement was observed in case of all three different dimensional cases of the Rosenbrock function, in case of COUNTER NICHING EA.

Table 5. Average fitness comparison for DGEA and COUNTER_NICHING_EA[*]. Dimension of each function in this case is 100. Both algorithms were executed till stagnation.

| Function | DGEA (Fixed Run) | DGEA (Stagnation) | C_EA* (Fixed Run) | C_EA* (Stagnation) |
|---|---|---|---|---|
| $f_{ack}(x)_{20D}$ | 8.05E-4 | 3.36e-5 | 1.08E-61 | 1.09E-62 |
| $f_{ack}(x)_{50D}$ | 4.61E-3 | 2.52E-4 | 1.01E-29 | 1.01E-30 |
| $f_{ack}(x)_{100D}$ | 0.01329 | 9.80E-4 | 1.00E-9 | 1.01E-10 |
| $f_{gri}(x)_{20D}$ | 7.02E-4 | 7.88E-8 | 4.6E-62 | 4.01E-62 |
| $f_{gri}(x)_{50D}$ | 4.40E-3 | 1.19E-3 | 1.01E-30 | 1.01E-31 |
| $f_{gri}(x)_{100D}$ | 0.01238 | 3.24E-3 | 1.80E-9 | 1.52E-10 |
| $f_{rtg}(x)_{20D}$ | 2.21E-5 | 3.37E-8 | 1.21E-61 | 1.00E-62 |
| $f_{rtg}(x)_{50D}$ | 0.01664 | 1.97E-6 | 2.01E-30 | 2.01E-31 |
| $f_{rtg}(x)_{100D}$ | 0.15665 | 6.56E-5 | 2.00E-9 | 2.00E-11 |
| $f_{ros}(x)_{20D}$ | 96.007 | 8.127 | 1.0E-60 | 1.0E-60 |
| $f_{ros}(x)_{50D}$ | 315.395 | 59.789 | 1.91E-29 | 1.90E-29 |
| $f_{ros}(x)_{100D}$ | 1161.550 | 880.324 | 3.00E-9 | 3.00E-9 |

### 6.3 An Analysis of Population Diversity for COUNTER NICHING EA

In the next phase of our experiments, we have investigated COUNTER NICHING EA's performance in terms of maintaining constructive diversity. There are various measures of diversity available. The "*distance-to-average-point*" measure used in [15] is relatively robust with respect to population size, dimensionality of problem and the search range of each variable. Hence, we have used this measure of diversity in our investigation. The "*distance-to-average-point*" measure for $N$ dimensional numerical problems can be described as below [15].

$$diversity(P) = \frac{1}{|L| \cdot |P|} \cdot \sum_{i=1}^{|P|} \sqrt{\sum_{j=1}^{N} \left( s_{ij} - \bar{s}_j \right)^2} \tag{1}$$

where, $|L|$ is the length of the diagonal or range in the search space $S \subseteq \Re^N$, $P$ is the population, $|P|$ is the population size, $N$ is the dimensionality of the problem, $s_{ij}$ is the $j$'th value of the $i$'th individual, and $\bar{s}_j$ is the $j$'th value of the average point $\bar{s}$. It is assumed that each search variable $s_k$ is in a finite range, $s_{k\_min} \leq s_k \leq s_{k\_max}$. Table 6 depicts the average diversity for four test problems with COUNTER NICHING EA simulation runs. The values reported in Table 6, averages the value of the diversity measure in equation (1) calculated at each generation where there has been an improvement in average fitness over 500, 1000 and 2000 generations for the 20, 50 and 100 dimensional cases respectively. Final values were averaged over 100 runs. To eliminate the noise in the initial generations of a run, diversity calculation does not start until the generation since which a relatively steady improvement in fitness has been observed. Table 6 shows that the COUNTER NICHING EA does not necessarily maintain very high average population diversity. However, EA's requirement is not to maintain very high average population diversity

but to maintain an optimal level of population diversity. The high solution accuracy obtained by COUNTER NICHING EA proves that the algorithm is successful in this respect.

Table 6. Average population diversity comparison for COUNTER NICHING EA (fixed run). An average of 100 runs have been reported in each case.

| Function | 20D | 50D | 100D |
|---|---|---|---|
| $f_{ack}(x)$ | 0.001350 | 0.001811 | 0.002001 |
| $f_{gri}(x)$ | 0.001290 | 0.001725 | 0.002099 |
| $f_{rtg}(x)$ | 0.003000 | 0.003550 | 0.004015 |
| $f_{ros}(x)$ | 0.001718 | 0.002025 | 0.002989 |

**6.4 Statistical Significance of Comparative Analysis**

Finally, a *t*-test (at 0.05 level of significance; 95% confidence) was applied in order to ascertain if differences in the "$A$" performance for the best average fitness function are statistically significant from the other techniques used for comparison. The $P$-values of the two-tailed *t*-test are given in Table 7. As can be observed, the difference in "$A$" performance of COUNTER NICHING EA is statistically significant compared to the majority of the techniques across the test functions in their three different dimensional versions.

Table 7. The $P$-values of the t-test with 99 degrees of freedom. Dimensions of each function considered are 20, 50 and 100. '-' appears where the corresponding data is not available.

| Function | $C\_EA^*$-SEA | $C\_EA^*$-SOCEA | $C\_EA^*$-CEA | $C\_EA^*$-DGEA |
|---|---|---|---|---|
| $f_{ack}(x)$ 20D | 0.1144 | 0.4263 | 0.625 | 0.9954 |
| $f_{gri}(x)$ 20D | 0.2793 | 0.3349 | 0.4231 | 0.9998 |
| $f_{rtg}(x)$ 20D | 0.0009 | 0.0901 | 0.2636 | 0.9999 |
| $f_{ros}(x)$ 20D | 0 | 0 | 0 | 0.0044 |
| $f_{ack}(x)$ 50D | 0.0903 | 0.217 | 0.4198 | 0.9873 |
| $f_{gri}(x)$ 50D | 0.2037 | 0.2843 | 0.3098 | 0.9725 |
| $f_{rtg}(x)$ 50D | 0 | 0 | 0.0002 | 0.9989 |
| $f_{ros}(x)$ 50D | 0 | 0 | 0 | 0 |
| $f_{ack}(x)$ 100D | 0.0891 | 0.1363 | 0.2857 | 0.975 |
| $f_{gri}(x)$ 100D | 0.1337 | 0.2019 | 0.2776 | 0.9546 |
| $f_{rtg}(x)$ 100D | 0 | 0 | 0 | 0 |
| $f_{ros}(x)$ 100D | 0 | 0 | 0 | 0 |

## 7  Conclusions

In this paper we investigated the issues related to population diversity in the context of the evolutionary search process. We established the association between population diversity and the search ability of a typical evolutionary algorithm. Then we presented an investigation on an intelligent mutation based EA that tries to achieve optimal diversity in the search landscape. The framework basically incorporates two key processes. *Firstly*, the population's spatial information is obtained with a pseudo-niching algorithm. *Secondly*, the information is used to identify potential local convergence and community formations. Then diversity is introduced with informed genetic operations, aiming at two objectives: (a) Promising samples from unexplored regions are introduced replacing *redundant* less fit members of over-populated communities and (b) While local entrapment is discouraged, representative members are still preserved to encourage *exploitation*. While the current focus of the research was to introduce and maintain population diversity to avoid local entrapment, this Counter Niching-based algorithm can also be adapted to serve as an inexpensive alternative for *niching* genetic algorithm, to identify multiple solutions in multimodal problems as well as to suit the diversity requirements in a dynamic environment.

## Reference

1. Adra, S. F. and Fleming, P. J. 2011. Diversity management in evolutionary many-objective optimization. IEEE Trans. Evol. Comput. 15, 2, 183–195.
2. Araujo, L. and Merelo, J. J. 2011. Diversity through multiculturality: Assessing migrant choice policies in an island model. IEEE Trans. Evol. Comput. 15, 4, 456–468.
3. Bhattacharya, M., "An Informed Operator Approach to Tackle Diversity Constraints in Evolutionary Search", Proceedings of The International Conference on Information Technology, ITCC 2004, Vol. 2, IEEE Computer Society Press, ISBN 0-7695-2108-8, 326-330.
4. Bhattacharya, M., "Counter-niching for Constructive Population Diversity", in Proceedings of the 2008 IEEE Congress on Evolutionary Computation (CEC 2008), Hong Kong, IEEE Press, ISBN: 978-1-4244-1823-7, 4174-4179.
5. Chow, C. K. and Yuen, S. Y. 2011. An evolutionary algorithm that makes decision based on the entire previous search history. IEEE Trans. Evol. Comput. 15, 6, 741–769.
6. Curran, D. and O'Riordan, C. 2006. Increasing population diversity through cultural learning. Adapt. Behav. 14, 4, 315–338.
7. Friedrich, T., Hebbinghaus, N., and Neumann, F. 2007. Rigorous analyses of simple diversity mechanisms. In Proceedings of the Genetic and Evolutionary Computation Conference. 1219–1225.

8. Friedrich, T., Oliveto, P. S., Sudholt, D., and Witt, C. 2008. Theoretical analysis of diversity mechanisms for global exploration. In Proceedings of the Genetic and Evolutionary Computation Conference. 945–952.
9. Ganv´an-L´opez, E., McDermott, J., O'Neill, M., and Brabazon, A. 2010. Towards an understanding of locality in genetic programming. In Proceedings of the 12th Annual Conference on Genetic and Evolutionary Computation. 901–908.
10. Gao, H. and Xu, W. 2011. Particle swarm algorithm with hybrid mutation strategy. Appl. Soft Comput. 11, 8, 5129–5142.
11. Jia, D., Zheng, G., and Khan, M. K. 2011. An effective memetic differential evolution algorithm based on chaotic local search. Inform. Sci. 181, 15, 3175–3187.
12. De Jong, K.A., An Analysis of the Behavior of a Class of Genetic Adaptive Systems, PhD thesis, University of Michigan, Ann Arbor, MI, Dissertation Abstracts International 36(10), 5140B, University Microfilms Number 76-9381, 1975.
13. Leung, Y., Gao, Y. and Xu, Z. B. 1997 Degree of Population Diversity-A Perspective on Premature Convergence in Genetic Algorithms and its Markov Chain Analysis ,IEEE Trans. Neural Networks. 8, 5, 1165-1176.
14. Liang, Y. and Leung, K. S. 2011. Genetic algorithm with adaptive elitist-population strategies for multimodal function optimization. Appl. Soft Comput. 11, 2, 2017–2034.
15. Ursem, R. K. 2002 Diversity-Guided Evolutionary Algorithms, Proceedings of Parallel Problem Solving from Nature VII (PPSN-2002),  462-471.
16. Thomsen, R., and Rickers, P. 2000. Introducing Spatial Agent-Based Models and Self-Organised Criticality to Evolutionary Algorithms. Master's thesis, University of Aarhus, Denmark.
17. B¨ack, T., Fogel, D. B., Michalewicz, Z. and others, (eds.). 1997. Handbook on Evolutionary Computation, IOP Publishing Ltd and Oxford University Press.
18. Bhattacharya, M. 2008. Meta Model Based EA for Complex Optimization. Int. J. Comp. Intell. 4.1.
19. Ishibuchi, H., Narukawa, K., Tsukamoto, N., and Nojima, Y. 2008. An empirical study on similarity-based mating for evolutionary multi-objective combinatorial optimization. Europ. J. Oper. Res. 188, 1, 57–75.
20. Bhattacharya, M. 2007Surrogate based EA for expensive optimization problems.  , IEEE Congress on Evolutionary Computation..
21. Bhattacharya, M. 2008Reduced computation for evolutionary optimization in noisy environment. Proceedings of the 10th annual Conference Companion on Genetic and Evolutionary Computation. ACM.
22. Bhattacharya, M. 2007 Expensive optimization, uncertain environment: an EA-based solution. Proceedings of the 2007 GECCO conference companion on Genetic and evolutionary computation. ACM.
23. Bhattacharya, M. 2008 Meta Model Based EA for Complex Optimization. International Journal of Computational Intelligence 4.1.
24. Bhattacharya, M. 2006Exploiting landscape information to avoid premature convergence in evolutionary search. IEEE Congress on Evolutionary Computation.
25. Bhattacharya, M. and Nath, B. 2001 Genetic programming: A review of some concerns. Computational Science-ICCS 2001. Springer Berlin Heidelberg. 1031-1040.